\documentclass{bmvc2k}

\usepackage{amsfonts}
\usepackage{booktabs}
\usepackage[ruled,vlined,linesnumbered]{algorithm2e}

\title{Dynamic Try-On: Taming Video Virtual Try-on with Dynamic Attention Mechanism}

\addauthor{Jun Zheng}{zhengj98@mail2.sysu.edu.cn}{1}
\addauthor{Jing Wang}{wangj977@mail2.sysu.edu.cn}{1}
\addauthor{Fuwei Zhao}{zhaofuwei.777@bytedance.com}{2}
\addauthor{Xujie Zhang}{zhangxj59@mail2.sysu.edu.cn}{1}
\addauthor{Xiaodan Liang}{xdliang328@gmail.com}{1,3}

\addinstitution{
 Shenzhen Campus of \\
 Sun Yat-sen University, China
}
\addinstitution{
 ByteDance, China
}
\addinstitution{
 Corresponding Author
}

\runninghead{Student, Prof, Collaborator}{BMVC Author Guidelines}

\begin{document}

\maketitle

\begin{figure*}[htbp]
\centering
    \includegraphics[width=0.9\linewidth]{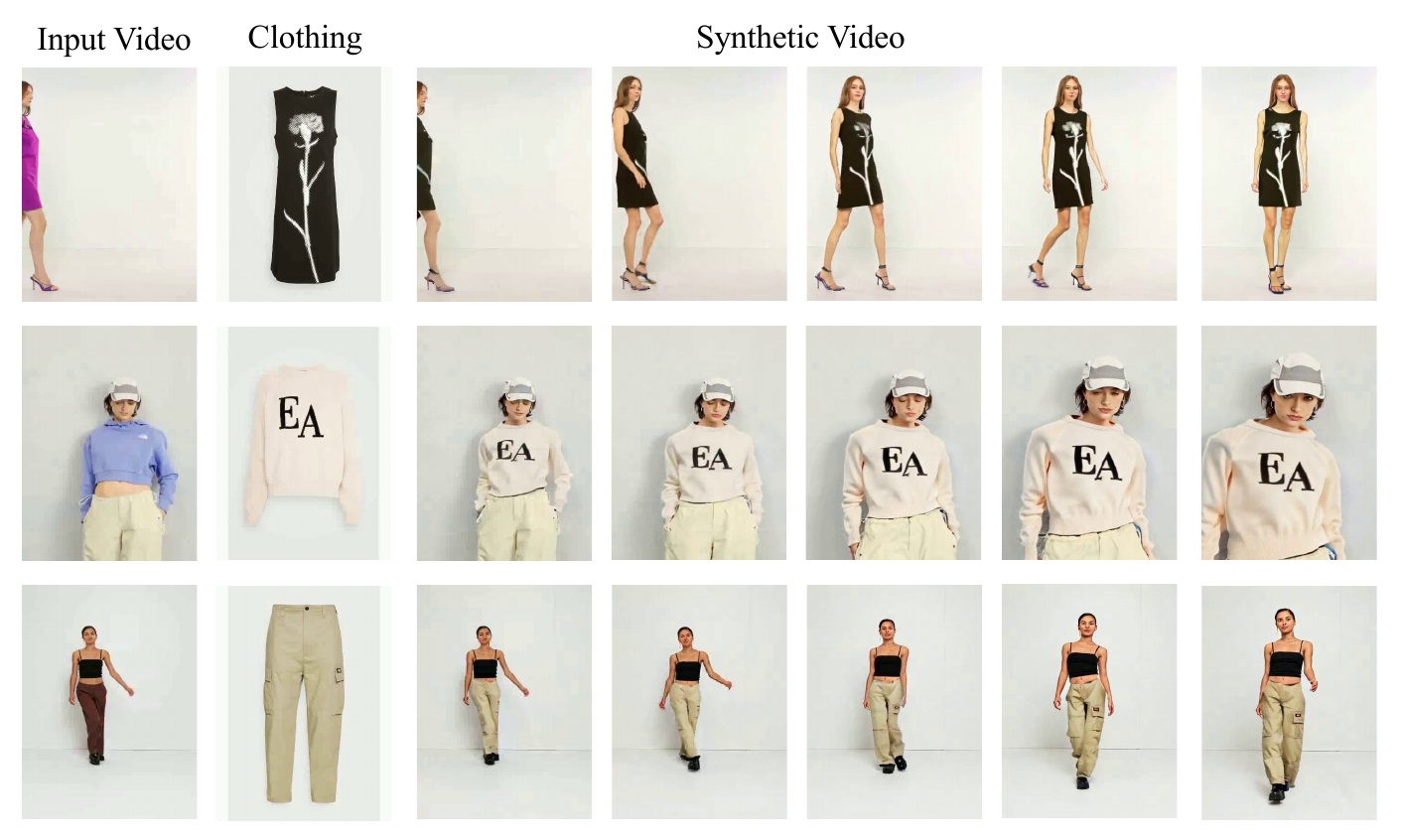}
   \caption{Video try-on results of the proposed Dynamic Try-On, illustrating its generalization capability across diverse clothing types.}
\label{fig:teaser}
\end{figure*}

\begin{abstract}
 Video virtual try-on is a promising research area with significant real-world applications. Previous research on video try-on has primarily focused on transferring product clothing images to videos with simple human poses, while performing poorly with complex movements. To better preserve clothing details, these approaches often employ an additional garment encoder, which increases computational resource consumption. The primary challenges in this domain are twofold: (1) leveraging the garment encoder's capabilities in video try-on while lowering computational requirements; (2) ensuring temporal consistency in the synthesis of human body parts, especially during rapid movements. To tackle these issues, we propose a novel video try-on framework based on Diffusion Transformer (DiT), named \textbf{Dynamic Try-On}. To reduce computational overhead, we repurpose the DiT backbone as the garment encoder and introduce a Dynamic Feature Fusion Module (DFFM) for efficient garment feature storage and integration. To enhance temporal consistency, particularly for human body parts, we introduce a Limb-aware Dynamic Attention Module (LDAM) that guides the DiT backbone to focus on limb regions during the denoising process. Extensive experiments demonstrate the superiority of Dynamic Try-On in generating stable and smooth try-on results, even for videos featuring complicated human postures.
\end{abstract}

\section{Introduction}
\label{sec:intro}

Video virtual try-on systems~\cite{fwgan, clothformer, xu2024tunnel, fang2024vivid,Karras_FashionVDM_2024} aim to seamlessly transfer desired clothing onto a target person in a video, while preserving their original motion and identity. This technology offers significant potential for applications in e-commerce. Although video representation offers a more compelling user experience, it presents greater technical challenges than image-based try-on. Therefore, the majority of existing work has focused on image-based try-on~\cite{XintongHan2018VITONAI,SeungHwanChoi2021VITONHDHV,liuLQWTcvpr16DeepFashion,YuyingGe2021ParserFreeVT,he2022fs_vton,NEURIPS2021_151de84c, xie2023gpvton, Xie2021TowardsSU}. The earlier approaches typically build on Generative Adversarial Networks (GANs)~\cite{NEURIPS2021_151de84c,he2022fs_vton,SeungHwanChoi2021VITONHDHV,xie2023gpvton, Xie2021TowardsSU}, containing a warping module and a try-on generator. The warping module deforms clothing to align with the human body, and then the warped garment is fused with the person image through the try-on generator. However, with the recent advent of UNet-based Latent Diffusion Models (LDMs)~\cite{ldm,controlnet,mou2023t2i,yang2022paint} and Transformer-based LDMs (or Diffusion Transformer, DiT)~\cite{stablediffusion3,sora,opensora,opensoraplan}, research attention has increasingly shifted towards these emerging generative models due to their potential for groundbreaking results. A diffusion-based try-on framework does not explicitly separate the warping and blending operations. Instead, it implicitly unifies them into a single cross-attention process facilitated by a specially designed powerful garment encoder. By utilizing text-to-image pre-trained weights, these diffusion approaches demonstrate superior fidelity compared to the GAN-based counterparts.

\begin{figure}[htbp]
    \centering
    \includegraphics[width=1.0\linewidth]{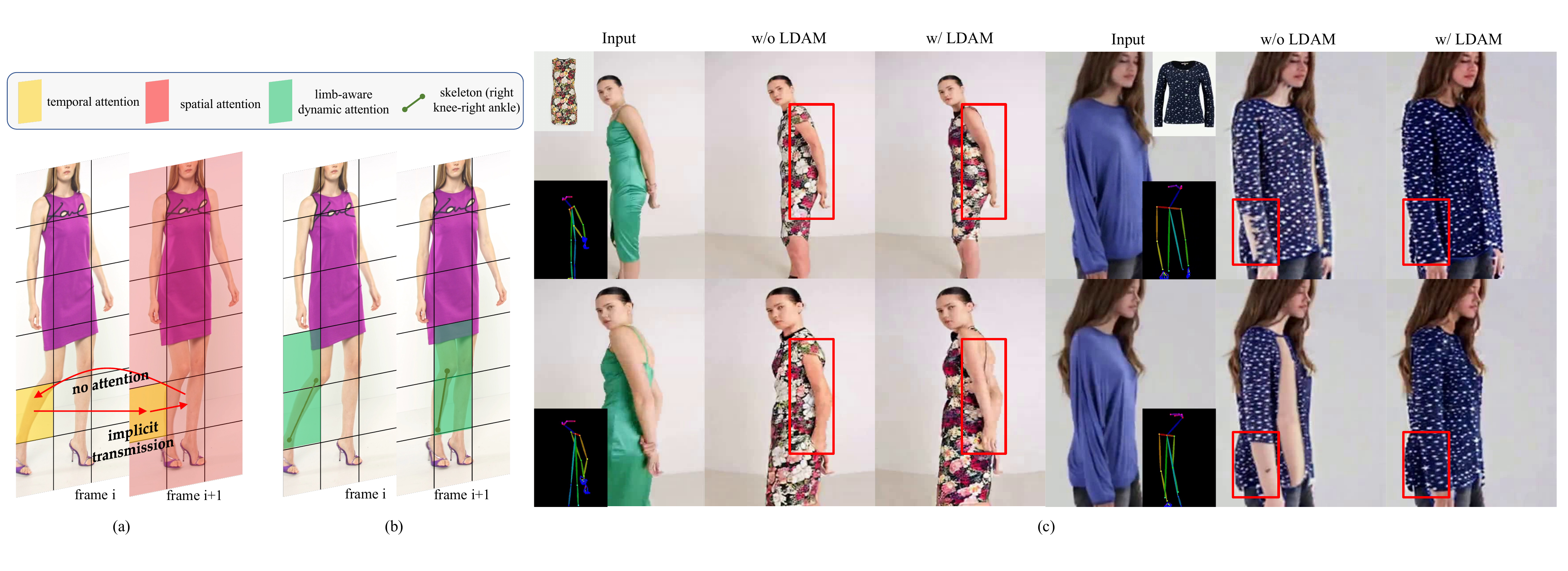}
    \caption{(a) The body part in frame $i+1$ cannot directly attend to the same part in frame $i$. Instead, body information can only be implicitly transmitted through other background patches. (b) Our limb-aware dynamic attention enables the model to effectively convey body information across frames. (c) Qualitative ablations for LDAM. It assists in generating appropriate human body parts, especially during rapid movements.}
    \label{fig:attn-compare}
\end{figure}

Recently, there are a few attempts of designing video try-on based on LDMs~\cite{xu2024tunnel, fang2024vivid, Karras_FashionVDM_2024}. These approaches typically rely on an extra garment encoder to produce visually pleasing try-on results, which significantly increases the VRAM consumption during model training. In terms of generating videos that align with the given human poses, prior methods~\cite{hu2023animateanyone,wang2024unianimate,xu2024tunnel} often employ compact pose encoders that may not enforce strict temporal coherence constraints. This limitation poses a challenge to developing robust video try-on frameworks. Before delving into the limitations of this design in detail, we provide some background information below. The popular paradigm for video LDMs~\cite{guo2023animatediff,2023i2vgenxl,2023videocomposer,xing2023dynamicrafter,guo2023sparsectrl,opensora} involves separate spatial and temporal attention modules. This separation facilitates the construction of video LDMs by building upon existing image LDMs through the insertion of temporal modules. However, as highlighted in CogVideoX~\cite{yang2024cogvideox}, the separation of spatial and temporal attention modules can make handling large inter-frame motions challenging. As illustrated in Fig.~\ref{fig:attn-compare}(a), this limitation sometimes leads to failure when these video try-on models encounter rapid movements in human videos. Consequently, due to the inherent limitations of this paradigm, many recent approaches, such as those described in~\cite{yang2024cogvideox, kong2024hunyuanvideo, ma2025stepvideot2vtechnicalreportpractice,xu2024easyanimatehighperformancelongvideo}, have shifted towards building more robust text-to-video models based on 3D full attention layers. While these newer models offer enhanced capabilities for motion handling, they are notably more computationally intensive.

To address the aforementioned challenges of VRAM consumption and rapid movement handling, we propose Dynamic Try-On, a novel DiT-based video try-on network. Our approach is designed for lower computational resource utilization while ensuring superior temporal consistency in human body part synthesis compared to existing methods. Fig.~\ref{fig:teaser} shows samples generated by our model\footnote{Please refer to the supplementary videos for more results.}. Specifically, Dynamic Try-On contains \textbf{D}ynamic \textbf{F}eature \textbf{F}usion \textbf{M}odule (\textbf{DFFM}) to preserve clothing details by storing and integrating garment features extracted by the DiT blocks and \textbf{L}imb-aware \textbf{D}ynamic \textbf{A}ttention \textbf{M}odule (\textbf{LDAM}). As shown in Fig.~\ref{fig:attn-compare}(b), by passing through LDAM, the limb-related tokens are selected and enforced to maintain temporally consistency. We refer to the combination of DFFM and LDAM as the \emph{dynamic attention mechanism} due to the dynamic operations within these modules.
To validate the performance of our framework, we collected an in-shop virtual try-on dataset with complicated human postures for our research purpose. Our experiments demonstrate that Dynamic Try-On outperforms the existing methods in generating videos, both quantitatively and qualitatively.

Our contributions can be summarized as follows: (1) We propose a novel DiT-based video try-on network, Dynamic Try-On, featuring consistent spatio-temporal generation on videos with complex human motions. (2) We propose dynamic feature fusion module to store and integrate garment features, enabling the precise recovery of clothing details in videos without the need for a bulky garment encoder. (3) We design limb-aware dynamic attention module to guarantee the temporal consistency of human body parts, surpassing 3D full attention layers in both VRAM consumption and performance.

\section{Related Work}
\label{sec:related_work}
\subsection{Video Virtual Try-on.} 
Existing work on video virtual try-on can be classified as GAN-based~\cite{fwgan,mv-ton,shineon, clothformer} and diffusion-based methods~\cite{xu2024tunnel, fang2024vivid, Karras_FashionVDM_2024}. The former relies on garment warping by optical flow~\cite{dosovitskiy2015flownet} and utilizes a GAN generator to fuse the warped clothing with the reference person. FW-GAN~\cite{fwgan} predicts optical flow to warp preceding frames during the video try-on process, thereby ensuring the generation of temporally coherent video sequences. ClothFormer~\cite{clothformer}  presents a dual-stream transformer architecture to efficiently integrate garment and person features, facilitating more accurate and realistic video try-on results. Despite achieving reasonable performance, GAN-based methods often struggle with garment-person misalignment, particularly when warping flow estimation is inaccurate. Moreover, their overall generation quality is often inferior to that of diffusion-based models, which benefit from large-scale pre-trained weights. The recent ViViD~\cite{fang2024vivid} proposes to use a UNet-based diffusion model for video try-on. It can handle camera movements and faithfully preserve the clothing textures. However, its demonstration videos primarily feature product images and simple human movements over limited frame counts. Instead, our Dynamic Try-On can be applied to videos with complicated postures and can generate long sequences with high-quality spatiotemporal consistency.

\subsection{Image Animation.}
Image animation aims to generate a video sequence from a static image. Recently, diffusion-based models have shown unprecedented success in this domain~\cite{hu2023animateanyone, xu2023magicanimate, wang2023disco, dreampose_2023}. Notably, MagicAnimate~\cite{xu2023magicanimate} has has demonstrated state-of-the-art generation results among open-source models. It utilizes an additional U-Net to extract appearance information from images and a pose encoder to process pose sequences.
Combining animation frameworks with image try-on methods can achieve video try-on, for example, basically apply image try-on methods to the first frame of frame sequences and then perform human animation. However, a significant drawback of this simplified pipeline is the potential absence of detailed garment information when a reference in-shop clothing image is unavailable, leading to unfaithful rendering of clothing details. Comparative experimental results for this two-stage pipeline, presented in Sec.~\ref{quant}, highlight the superiority of our integrated Dynamic Try-On approach.

\section{Method}
\label{method}
We present Dynamic Try-On, a video virtual try-on framework built upon diffusion transformers (DiT)~\cite{Peebles2022DiT}. 
Before introducing our architecture, we briefly review basic concepts of Latent Diffusion Models and DiT in Sec.~\ref{preliminary}. The overall architecture will be presented in Sec.~\ref{arch}, and we introduce details of DFFM and LDAM in Sec.~\ref{DFFM} and Sec.~\ref{LDAM}.

\begin{figure*}[htbp]
    \centering
    \includegraphics[width=0.95\linewidth]{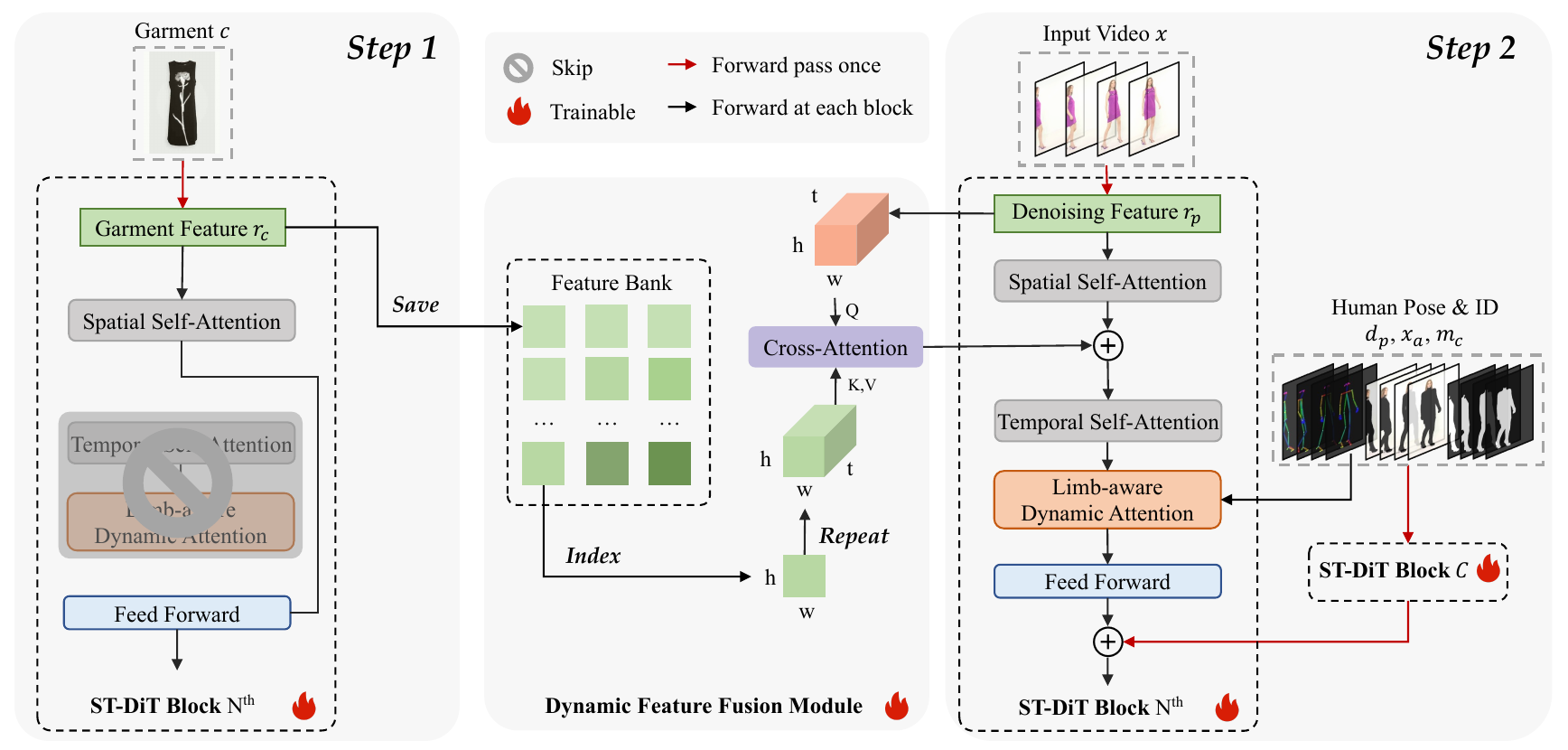}
    \caption{Overview of the proposed Dynamic Try-On. \textit{Step 1}: extracting garment features via a chain of blocks.  \textit{Step 2}: delivering garment features and injecting human pose information into blocks, thus generating high-quality try-on videos.}
    \label{fig:framework}
\end{figure*}


\subsection{Preliminary}
\label{preliminary}

\subsubsection{Latent Diffusion Models (LDMs)} Generating high-resolution images/videos directly in the original pixel space can be computationally expensive and challenging due to the high dimensionality. Instead, LDMs~\cite{ldm} operate in a latent space where the data is represented in a more compact form. This approach leverages the power of variational autoencoders (VAEs)~\cite{vae} to encode the high-dimensional data into a latent space and then apply the diffusion process in this latent space. An image LDM typically contains three key components: (a) an Encoder $\mathcal{E}$ mapping the high-resolution image $x$ to a latent representation $z=\mathcal{E}\left(x\right)$\,, (b) a Diffusion Process involving a forward process that gradually adds noise to $z$ over $T$ time steps:
    $q\left(z_t \mid z_{t-1}\right)=\mathcal{N}\left(z_t ; \sqrt{1-\beta_t} z_{t-1}, \beta_t I\right)$, where $\beta_{t}$ is a variance schedule that controls the amount of noise added at each step; and a reverse process parameterized by a neural network (typically a U-Net~\cite{unet}) $p_{\theta}$ that learns to denoise: 
    $p_\theta\left(z_{t-1} \mid z_t\right)=\mathcal{N}\left(z_{t-1} ; \mu_\theta\left(z_t, t\right), \sigma_\theta^2(t) I\right)$,
 (c) a Decoder $\mathcal{D}$ maps the denoised latent representation back to the original image space: $\hat{x}=\mathcal{D}_{\left(z_0\right)}$. The training objective is typically a reconstruction loss in the latent space that minimizes the noise $\epsilon$ and the network's prediction: 
    $L_{L D M}=\mathbb{E}_{z, \epsilon, t}\left[\left\|\epsilon-p_\theta\left(z_t, t\right)\right\|_2^2\right]$.
Once Trained, we can sample $z_t$ from $p_{(z)}$ and decode it to image space with a single pass through $\mathcal{D}$.

\subsubsection{Diffusion Transformers (DiT)} The Diffusion Transformer~\cite{Peebles2022DiT} is an innovative architecture that leverages the strengths of diffusion models and transformers~\cite{Vaswani2017AttentionIA}. By integrating these two powerful paradigms, it aims to enhance the quality, flexibility, and scalability compared to traditional UNet-based LDMs \cite{ldm}. The overall formulation remains the same as the LDMs except using a transformer (instead of a UNet) to learn the denoising function $p_{\theta}$ within a diffusion-based framework. To fully leverage our dynamic attention mechanism, we adopt a modified Spatio-Temporal DiT (ST-DiT) as the backbone of our Dynamic Try-On. 
\subsection{Overall Architecture}
\label{arch}
This section provides a comprehensive illustration of the pipeline presented in Fig.~\ref{fig:framework}. We start with introducing the formulation of video try-on task. Afterwards, we briefly describe our novel dynamic attention mechanism which will be elaborated on in the next sections.

\subsubsection{Formulation of Video Try-On Task}
Video virtual try-on can be viewed as a video inpainting problem. It requires a four-tuple $\left\{x_a, d_p, m_c, c\right\}$ to place the target clothing $c$ on the reference person video $x$, including the cloth-agnostic frame $x_a$, the pose skeleton frame $d_p$ and the inpainting mask frame $m_c$, as visualized in Fig.~\ref{fig:framework}. As the pre-trained weights are not tuned for inpainting, we introduce a ST-DiT block $\mathcal{C}$ to preserve the person's pose, identity and background. Specifically, $\mathcal{C}$ is a trainable replica of the first block of the denoising backbone. We add the output of $\mathcal{C}$ as residual solely to the first block of the denoising DiT. 

\subsubsection{Step 1: Garment Feature Extraction}
\label{step1} As shown in Fig.~\ref{fig:framework} step 1, the garment image $c$ is encoded by $\mathcal{E}$ and then passes through $N$ ST-DiT blocks (with $\mathcal{E}$ omitted in the figure for simplicity). The intermediate garment features are stored in the feature bank. During this procedure, both temporal self-attention and limb-aware dynamic attention are skipped, as the processing involves only a single garment image without human pose information.

\subsubsection{Step 2: Dynamic Attention Mechanism}
\label{DAM} The dynamic attention mechanism comprises two key components: the Dynamic Feature Fusion Module (DFFM) and the Limb-aware Dynamic Attention Module (LDAM). As described in Sec.~\ref{step1}, garment features have been stored in the feature bank of DFFM during step 1. As illustrated in Step 2 of Fig.~\ref{fig:framework}, when the denoising feature passes through each ST-DiT block, the corresponding garment feature is retrieved from the feature bank and fused with the denoising feature via DFFM. Simultaneously, the human pose sequence is utilized as prior knowledge to enhance the denoising feature within the LDAM.

\subsection{Dynamic Feature Fusion Module}
\label{DFFM}
Accurately recovering the texture details of desired garments is crucial for high-quality video try-on results. To this end, prior approaches~\cite{zheng2024vitondit, xu2024tunnel, fang2024vivid, Karras_FashionVDM_2024} adopt a garment encoder in parallel with the backbone. There are two main variations of this design. Fashion-VDM~\cite{Karras_FashionVDM_2024} uses a replica of the front half of the backbone as the garment encoder, while ViViD~\cite{fang2024vivid} and Tunnel Try-on~\cite{xu2024tunnel} directly use a copy of the entire backbone. In contrast to these common garment preservation paradigms, our DFFM offers a lightweight yet effective replacement, while possessing capabilities on par with them. Briefly, the functionality of DFFM involves two forward passes through the backbone. In the first pass (step 1), garment features are extracted by blocks and stored in the feature bank of DFFM. In the second pass (step 2), the denoising feature is combined with the stored garment feature through an additive attention process, facilitating the seamless integration of clothing characteristics into the video generation process. Here, we provide a more detailed formulation of the process.

Before passing through the backbone, the input video $x \in \mathbb{R}^{f \times H \times W \times 3}$ is first projected into the latent space, producing the latent $z_0 \in \mathbb{R}^{f \times h \times w \times 4} = \mathcal{E}\left(x\right)$\,, where $h=H/8$, $w=W/8$, and $f$ refers to the number of frames. Given patch size $p \times p$, the spatial represented $z_0$ is noised to produce $z_t$ and then ``patchified'' into a sequence of length $s = hw / p^2$ with hidden dimension $d$, forming the denoising feature $r_p \in \mathbb{R}^{f \times s \times d}$. Similarly, we can formulate the intermediate garment feature as $r_c\in \mathbb{R}^{1 \times s \times d}$ without adding noise.
As shown in Fig.~\ref{fig:framework} step 2, when the denoising feature $r_p$ passing through the denoising DiT, the corresponding garment feature $r_c$ will be retrieved from the feature bank and also duplicated $f$ times to match the shape of $r_p$. Lastly, $r_p$ and $r_c$ are run through the cross-attention in DFFM and the output features are added back to $r_p$ as a residual connection.
A line of work~\cite{hu2023animateanyone,wang2024stablegarment,xu2024ootdiffusion,zheng2024vitondit, xu2024tunnel, fang2024vivid, Karras_FashionVDM_2024} has proved the effectiveness of this attention fusion operation in keeping the texture details. We differ from them in our newly inserted cross-attention layer and the reusable backbone. By using an additional cross-attention layer and directly utilizing the denoising backbone itself as the garment encoder, we improve the model's capacity and capability to perceive the garment features while decreasing computational demands. While DFFM ensures accurate garment feature integration efficiently, maintaining the temporal coherence of articulated body parts, especially during rapid motion, requires dedicated handling, which motivates our second component: the Limb-aware Dynamic Attention Module.

\vspace{-3mm}
\subsection{Limb-aware Dynamic Attention Module}
\label{LDAM}
\begin{figure*}[h!]
    \centering
    \includegraphics[width=0.95\linewidth]{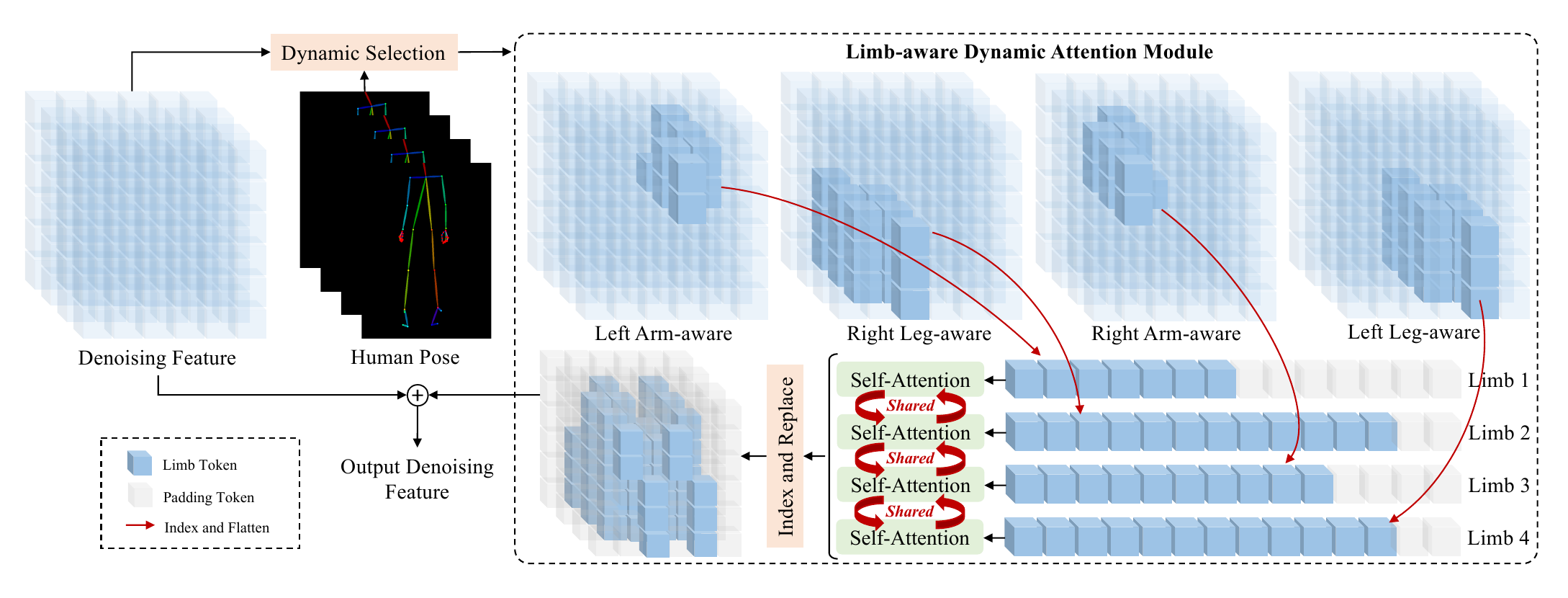}
    \caption{Visualization of Limb-aware Dynamic Attention Module}
    \label{fig:limb-attn}
\end{figure*}

With rapid movements of the human body, a basic combination of spatial and temporal attention struggles to maintain the temporal consistency of the limbs, especially when they overlap or temporarily move out of view. Meanwhile, as shown in Tab.~\ref{tab:attn-comp}, the computational complexity of 3D full attention~\cite{yang2024cogvideox, xu2024easyanimatehighperformancelongvideo,kong2024hunyuanvideo, ma2025stepvideot2vtechnicalreportpractice}, crucial for enhancing both temporal and spatial consistency, becomes overwhelming. To balance efficiency and performance, we propose the \textbf{L}imb-aware \textbf{D}ynamic \textbf{A}ttention \textbf{M}odule (LDAM), based on the given human pose sequence (e.g., keypoint locations mapped to token indices), dynamically indexes, groups, and models the tokens of different limbs from the person denoising feature, ensuring the consistency of each limb throughout the entire generated video, as shown in Fig.~\ref{fig:limb-attn}.

\begin{table}[]
    \centering
\resizebox{0.99\linewidth}{!}{
\begin{tabular}{@{}lcccc@{}}
\toprule
Attention Type   & Input Tensor Shape & Sequence Length & Complexity  & Explanation\\ \midrule
Spatial Attention &  $[B \times f, s, d]$    & $s$     & $\mathcal{O}(Bfs^2d)$  & the number of spatial token $s$    \\
Temporal Attention&   $[B \times s, f, d]$    & $f$   &$ \mathcal{O}(Bsf^2d)$ & the number of temporal token $T$    \\ 
3D Full Attention &   $[B, f \times s, d]$    & $f\times s$   &$ \mathcal{O}(B(fs)^2d)$  & batch size $B$ \\
LDAM     &   $[B \times L, n, d]$    & $n$   &$ \mathcal{O}(BLn^2d)$  & the number of limbs $L$, token length $n$  \\ \bottomrule
\end{tabular}}
    \caption{Theoretical comparison of different attention types. It is important to note that in this comparison, the number of limbs $L=4$ and during training at a resolution of $192 \times 256$, the parameter $n=12 \ll f \times s=192\times 36$.}
    \label{tab:attn-comp}
\end{table}

Specifically, given the denoising feature $r_p \in \mathbb{R}^{f \times s \times d}$ and the human limb token mask $S_l \in \{0, 1\}^{L \times f \times s }$ (including $L$ limbs, i.e. left arm, right arm, etc.), we first retrieve the corresponding limb features from $r_p$ according to $S_l$, and then align their spatial dimension to the same length $n$ by padding tokens. Here we obtain the limb features $r_l \in \mathbb{R}^{L \times n \times d}$ and simultaneously get the attention mask $M_l \in \mathbb{R}^{L \times n \times n}$ ready for masked self-attention calculation~\cite{stablepose2024}. Next, we compute masked self-attention for limb feature $r_l$ to obtain $r_l' \in \mathbb{R}^{L \times n \times d}$. We pass $r_l'$ through a zero-initialized linear layer, and then we add $r_l'$ back to $r_p$ according to the index $S_l$. Through the above process, we develop a plug-and-play LDAM that complements spatial and temporal attention, offering a flexible and efficient solution for enhancing the temporal consistency of the human body. See the supplementary materials for more details.

\section{Experiments}
\vspace{-2mm}
\subsection{Datasets}
We conduct an evaluation of our Dynamic Try-On using two video try-on datasets: the VVT dataset~\cite{fwgan} and a custom-collected dataset. The VVT dataset serves as a conventional video virtual try-on dataset, including 791 paired person videos and clothing images with a resolution of $192\times 256$. The train and test set contain 159,170 and 30,931 frames, respectively. 
To better assess the performance of our method under complex human poses and occlusions, we curated an in-shop video dataset from an e-commerce platform. Our custom dataset contains 9,100 video-image pairs. For training and evaluation purposes, it is split into 9,000 videos (504,215 frames) for training and 100 videos (5,321 frames) for testing.
\vspace{-3mm}
\subsection{Implementation Details}
\label{sec:impl}
\textbf{Multi-Stage Training.}
During training, we progressively train the model. We first load pre-trained weights of OpenSora~\cite{opensora}, and organize the training into three distinct stages. In the first stage, we only train spatial self-attention and cross-attention layers to reconstruct the person image with corresponding in-shop garment image. Then we incorporate the ST-DiT block $\mathcal{C}$ and set all parameters trainable for the second stage. In the third stage, we plug in LDAM to the denoising backbone and train only the newly added modules.

\noindent\textbf{Hyper-parameters Setting.} We train Dynamic Try-On with two resolutions of $192\times 256$ (VVT dataset) and $384\times 512$ (our dataset). We use the $192\times 256$ version for a qualitative and quantitative comparison with baselines on the standard VVT dataset~\cite{fwgan}. And the $384\times 512$ version is for demo purposes. We adopt the AdamW optimizer~\cite{adamw} with a fixed learning rate of $1\times 10^{-5}$. The models are trained on 8 A100 GPUs. In the first stage, we utilized paired image data extracted from video datasets, and merged them with the existing VITON-HD dataset~\cite{SeungHwanChoi2021VITONHDHV}. Please check supplementary materials for more details.

\begin{figure*}[htbp]
    \centering
    \includegraphics[width=0.95\linewidth]{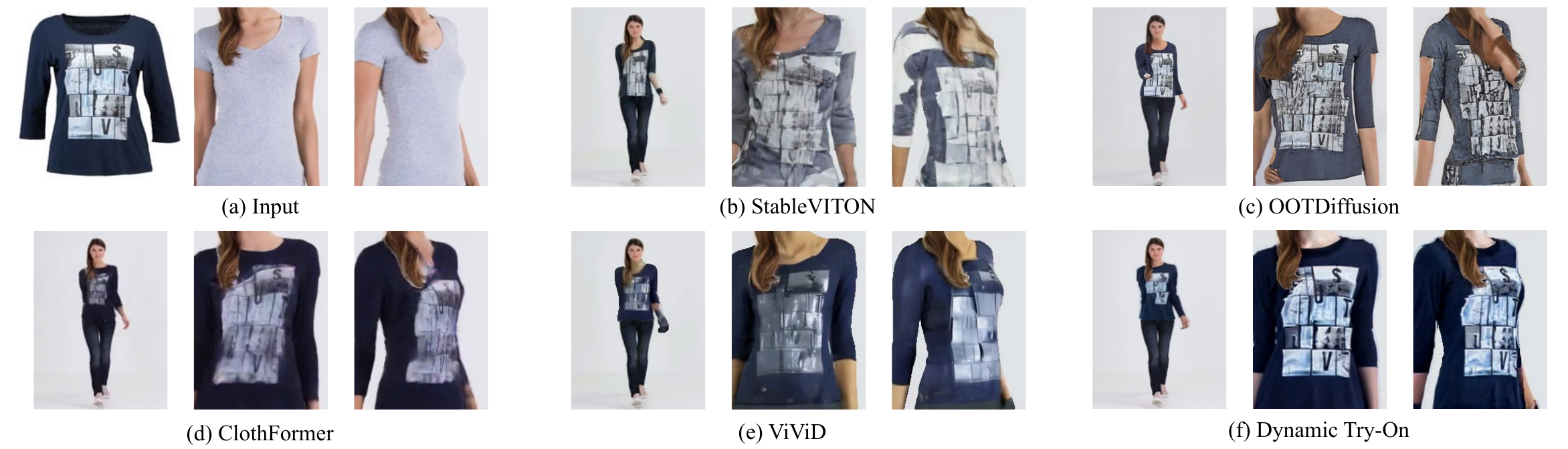}
    \caption{Qualitative comparison with baselines.}
    \vspace{-5mm}
    \label{fig:qualitative}
\end{figure*}

\vspace{-3mm}
\subsection{Qualitative Results}
Fig.\ref{fig:qualitative} presents the visual comparison between Dynamic Try-On and other baselines on the VVT dataset. It is clear that GAN-based ClothFormer~\cite{clothformer} (Fig.~\ref{fig:qualitative}(d)), is prone to clothing-person misalignment due to the inaccurate garment warping procedure. Although ClothFormer can handle smaller proportions of people, the generated images are often blurry and exhibit distorted cloth texture. Diffusion-based methods such as StableVITON~\cite{kim2023stableviton} and OOTDiffusion~\cite{xu2024ootdiffusion} produce relatively accurate single frame results for the full-body pose but fail for extremely close viewpoint. Furthermore, due to the image-based training, StableVITON and OOTDiffusion do not account for temporal coherence, resulting in noticeable jitters between consecutive frames (Fig.~\ref{fig:qualitative}(b) and Fig.~\ref{fig:qualitative}(c)). ViViD~\cite{fang2024vivid} in Fig.~\ref{fig:qualitative}(e) is a concurrent work that adapts U-Net diffusion model to video try-on. Despite reasonable results, the generated clothes exhibit obvious texture discrepancy compared with the ground truth.

In contrast, our Dynamic Try-On seamlessly integrates DFFM to the denoising DiT, allowing for accurate single-frame try-on with high inter-frame consistency. As depicted in Fig.~\ref{fig:qualitative}(f), the letters on the chest of the clothing adhere to the input shape and color, and are correctly positioned as the subject moves closer to the camera.
Furthermore, we provide additional qualitative results using our newly collected dataset to demonstrate the robust try-on capabilities and practicality of our Dynamic Try-On. Fig.~\ref{fig:teaser} showcases various results generated by Dynamic Try-On, including garment with special textures and scenarios involving complex motions. 
\vspace{-4mm}
\subsection{Quantitative Results}
\label{quant}
Quantitative results are reported in Tab.~\ref{tab:quant}. We adopt Structural Similarity Index (SSIM) ~\cite{2004SSIM} and Learned Perceptual Image Patch Similarity (LPIPS)~\cite{zhang2018unreasonable} as the frame-wise evaluation metrics. To assess the video-based performance, we concatenate every consecutive 10 frames to form a sample, and employ Video Fréchet Inception Distance (VFID)~\cite{fwgan} and Fréchet Video Distance (FVD)~\cite{Unterthiner2018TowardsAGFVD} that utilizes 3D convolution networks~\cite{i3d_metric} to evaluate both the visual quality and temporal consistency of the generated results. 
For the image-based evaluation, we compare our method with PBAFN~\cite{YuyingGe2021ParserFreeVT}, StableVITON~\cite{kim2023stableviton}, and OOTDiffusion~\cite{xu2024ootdiffusion}. For the video-based evaluation, we compare our method with FW-GAN~\cite{fwgan}, ClothFormer~\cite{clothformer}, Tunnel Try-on~\cite{xu2024tunnel} and ViViD~\cite{fang2024vivid}. Additionally, we use StableVITON~\cite{kim2023stableviton} and OOTDiffusion~\cite{xu2024ootdiffusion} combined with MagicAnimate\cite{chen2024magic} as the video baselines.

It is clear that Dynamic Try-On outperforms other methods, highlighting the advantages of our specially designed DFFM and LDAM. As shown in the top half of Table~\ref{tab:quant}, image-based methods struggle to achieve low video scores, indicating their limitations in modeling inter-frame consistency. Additionally, two-stage pipelines ("StableVITON + MA" and "OOTDiffusion + MA") perform worse than end-to-end approaches (ViViD~\cite{fang2024vivid} and Dynamic Try-On) even in video metrics, suggesting the limited capacity of current human animation methods when applied in video try-on scenarios.

\begin{table}
\parbox{.49\linewidth}{
    \centering
    \resizebox{0.99\linewidth}{!}{
    \begin{tabular}{@{}lcccc@{}}
    \toprule
    Method        & SSIM$\uparrow$   & LPIPS$\downarrow$ & VFID$\downarrow$ & FVD$\downarrow$ \\ 
    \midrule
    PBAFN~\cite{YuyingGe2021ParserFreeVT}         & 0.870          & 0.157 & 4.516     & -            \\
    StableVITON~\cite{kim2023stableviton}  & 0.914          & 0.132 & 6.291          & 220.05     \\
    OOTDiffusion~\cite{chen2023anydoor}     & 0.863          & 0.154 & 7.852  & 205.03  \\
    \midrule
    FW-GAN~\cite{fwgan}        & 0.675          & 0.283 & 8.019           & -   \\
    ClothFormer~\cite{clothformer}   & 0.921 & 0.081 & 3.967         & -         \\
    Tunnel Try-on~\cite{xu2024tunnel}   & 0.913 & \textbf{0.054} & 3.345         & -         \\
    StableVITON + MA  & 0.888          & 0.145 & 3.655   & 66.24            \\
    OOTDiffusion + MA       & 0.851          & 0.159 & 4.465   & 89.17             \\
    ViViD~\cite{fang2024vivid}& 0.913          & 0.133 & 2.961  & 66.14            \\
    \textbf{Dynamic Try-On}     & \textbf{0.924}          & 0.098 & \textbf{2.246}       & \textbf{57.49}           \\ 
    \bottomrule
    \end{tabular}}
        \caption{Quantitative comparison on VVT dataset. "MA" is short for MagicAnimate. The best results are denoted as \textbf{Bold}.}
    \label{tab:quant}}
\hfill
\parbox{.49\linewidth}{
    \centering
    \resizebox{0.99\linewidth}{!}{
\begin{tabular}{@{}lcccc@{}}
\toprule
Method          & 20 blocks & 28 blocks & 36 blocks & 44 blocks \\ \midrule
w/o DFFM &  42.6G     & 58.3G     & 74.0G     & OOM       \\
w/ DFFM            &  35.4G     & 48.5G     & 61.3G     & 74.2G     \\ \bottomrule
\end{tabular}}
    \caption{Quantitative comparison of different garment preservation paradigm regarding training memory cost (GB). “OOM” is short for out of memory. Note that our DFFM saves GPU memory, especially when the number of backbone blocks increases.}
    \label{tab:dffm}
    }
\end{table}
\subsection{Ablation Study}
To verify the effectiveness of DFFM and LDAM, we conduct two ablation experiments.

\noindent
\textbf{Effect of Dynamic Feature Fusion Module (DFFM).} As mentioned in Sec.~\ref{DFFM}, DFFM can notably reduce the computational resource requirements compared to previous garment preservation paradigm. Tab.~\ref{tab:dffm} shows our investigation into the effectiveness of DFFM using the same training settings for $192\times 256$ model. As the model size increases, the training memory required by the previous method grows more rapidly than that of DFFM due to the additional trainable parameters. Furthermore, comparable quantitative results in Tab.~\ref{tab:ablation} demonstrate DFFM's capability to preserve clothing details effectively.

\begin{table}[]
    \centering
\resizebox{0.99\linewidth}{!}{
\begin{tabular}{cc|ccc|cc|cccc}
\toprule
\multicolumn{2}{c|}{Garment Preservation Paradigm} & \multicolumn{3}{c|}{Additional Attention Layers} & \multicolumn{2}{c|}{\shortstack{Training Memory Cost of \\ Different Resolutions}} & \multicolumn{4}{c}{Evaluation Metrics} \\ \midrule
w/o DFFM & w/ DFFM  & None  & 3D Full Attention & LDAM & $192 \times 256$ & $384\times 512$ & SSIM$\uparrow$   & LPIPS$\downarrow$ & VFID$\downarrow$ & FVD$\downarrow$ \\ \midrule
\checkmark &     &  \checkmark&   &  & 58.3G & OOM &  0.918 & \textbf{0.092} & 2.493 & 63.53\\
 & \checkmark  &  \checkmark&     & & 48.5G & 73.8G    &  0.915 & 0.104 & 2.487 & 66.25\\
 & \checkmark &  & \checkmark  &   & 78.6G & OOM  &  0.918 & 0.105 & 2.451 & 60.73\\
  &  \checkmark&  &     & \checkmark& 57.6G & 79.3G &  \textbf{0.924} & 0.098 & \textbf{2.246} & \textbf{57.49}\\
\bottomrule
\end{tabular}}
    \caption{Ablation study of our proposed DFFM and LDAM on VVT dataset. "OOM" is short for out of memory.}
\label{tab:ablation}
\end{table}


\noindent
\textbf{Effect of Limb-aware Dynamic Attention Module (LDAM).} Considering LDAM as a form of sparse 3D attention, we compare it with 3D full attention layers to highlight its advantages. Due to the high computational cost of 3D full attention, we only insert 3D full attention layers into the first seven DiT blocks, thereby preventing out-of-memory issues. As demonstrated in Tab.~\ref{tab:ablation} and Fig.~\ref{fig:attn-compare}(c), the introduction of LDAM yields significant performance improvements across all evaluated metrics, even better than 3D full attention. In Fig.~\ref{fig:attn-compare}(c), the red-boxed area illustrates that without LDAM, the model struggles with overlapping limbs, often resulting in erroneous outputs. Conversely, LDAM effectively addresses these challenges, producing more coherent and accurate results.

\vspace{-4mm}
\section{Conclusions}
In this paper, we propose Dynamic Try-On, an innovative DiT-based video try-on framework that introduces novel designs to existing attention modules. By utilizing dynamic attention mechanism, Dynamic Try-On faithfully recovers clothing details and guarantees consistent body movements in the generated videos. Experiments highlight Dynamic Try-On's capability to handle diverse clothing and complex body movements, outperforming previous methods in all aspects.


\section{Acknowledgements}
This work is supported by Shenzhen Science and Technology Program \\ No.GJHZ20220913142600001, Nansha Key R\&D Program under Grant No.2022ZD014 and General Embodied AI Center of Sun Yat-sen University. This work is also sponsored by Doubao Fund.

\bibliography{egbib}

\end{document}